\pdfoutput=1
\documentclass[conference]{IEEEtran}
\usepackage[nocompress]{cite}
\usepackage{amsmath,amssymb,amsfonts}
\usepackage{algorithmic}
\usepackage{algorithm}
\usepackage{graphicx}
\usepackage{textcomp}
\usepackage{xcolor}
\usepackage{booktabs}
\usepackage{tabularx}
\usepackage{multirow}
\usepackage{hyperref}
\usepackage{subcaption}

\makeatletter
\def\abstract{\normalfont
    \if@twocolumn
      \@IEEEabskeysecsize\bfseries\textit{\abstractname}:\ \relax
    \else
      \bgroup\par\addvspace{0.5\baselineskip}\centering\vspace{-1.78ex}\@IEEEabskeysecsize\textbf{\abstractname}\par\addvspace{0.5\baselineskip}\egroup\quotation\@IEEEabskeysecsize
    \fi\@IEEEgobbleleadPARNLSP}
\def\IEEEkeywords{\normalfont
    \if@twocolumn
      \@IEEEabskeysecsize\bfseries\textit{\IEEEkeywordsname}:\ \relax
    \else
      \bgroup\par\addvspace{0.5\baselineskip}\centering\@IEEEabskeysecsize\textbf{\IEEEkeywordsname}\par\addvspace{0.5\baselineskip}\egroup\quotation\@IEEEabskeysecsize
    \fi\@IEEEgobbleleadPARNLSP}
\makeatother

\graphicspath{{figures/}}
\begin{document}

\title{Calibrate, Then Route: A Measured Study of Learned Request Routing\\ for Disaggregated LLM Serving}

\author{
\IEEEauthorblockN{Srikanta Datta Tumkur, Jay Iyer, Mehar Simhadri,\\ Sai Pavan Kumar, Sai Kapil Kumar, Ramesh Nampelly}
\IEEEauthorblockA{Vizuara}
}

\maketitle

\begin{abstract}
Disaggregated LLM serving separates the compute-bound prefill phase from the memory-bound decode phase onto distinct GPU pools, and systems such as DistServe, Splitwise, and Mooncake have made the mechanism fast. The mechanism does not decide placement, however. Every request still needs a prefill instance and a decode instance, and production systems make that choice with policies that are blind to the request itself. We study the routing policy directly. Our router scores each candidate instance by the marginal completion time a request would experience there, computed from four admission-time features: exact prompt length, predicted output length, post-admission KV-cache pressure, and SLO class. We first develop the policy in a discrete-event simulator, then validate it end to end on a real disaggregated cluster: eight NVIDIA A40 GPUs running one vLLM engine each, KV caches moved between pools by NIXL, and every workload driven at a measured saturation point. Across three arrival traces of mixed bursty traffic the calibrated learned router attains the highest mean goodput (0.864 versus 0.835 to 0.847 for round-robin, least-loaded, and a length heuristic) with the smallest variance across traces. It leads round-robin and the length heuristic on all three traces; against least-loaded it leads on two of three and on the third trails it by $0.003$, within run-to-run noise. Two further results qualify this finding. First, the router's cost constants have to be fitted to the target hardware. The identical policy running with simulator-derived constants loses 4.5 goodput points and roughly 40\% of its tail-latency advantage, because a mispriced cost model collapses the scorer into queue counting. Second, the router's advantage depends on where it operates. It grows with decode-pool width and traffic heterogeneity, vanishes on three-instance pools where a queue count is nearly a sufficient statistic, and inverts under extreme scarcity, where greedy cost minimization concentrates requests on whichever instance prices cheapest and blind spreading performs better. The learned router reaches with six GPUs the goodput that round-robin achieves with seven.
\end{abstract}


\section{Introduction}

\begin{figure*}[t]
\centering
\includegraphics[width=0.98\textwidth]{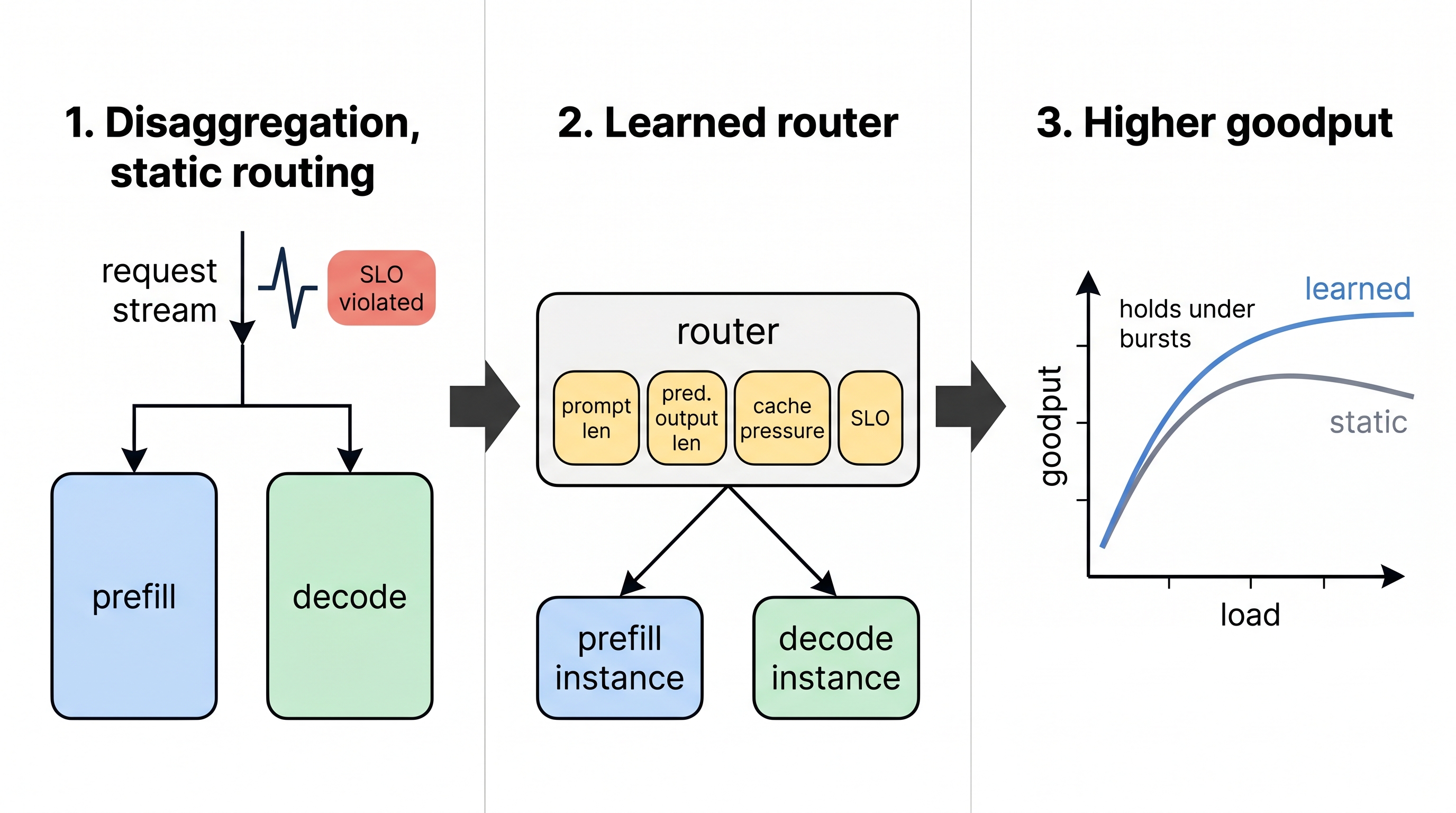}
\caption{Study overview. Disaggregation removes prefill/decode interference but leaves a per-request placement decision that is usually made blindly. A learned router prices each candidate instance from admission-time request features and is validated here on a real vLLM/NIXL cluster.}
\label{fig:overview}
\end{figure*}

An LLM inference request combines two different jobs. The first job reads the entire prompt in one compute-saturating pass; the second emits output tokens one at a time while its KV cache grows in GPU memory. Co-locating the two on one GPU causes them to interfere \cite{agrawal2024sarathi,kwon2023vllm}, so a family of serving systems now runs them on separate pools with an explicit KV-cache handoff in between: DistServe \cite{zhong2024distserve}, Splitwise \cite{patel2024splitwise}, and Mooncake \cite{qin2024mooncake}.

Splitting the phases creates a decision that the split itself does not resolve. Each arriving request must be matched to one prefill instance and one decode instance, and the quality of that match is invisible to policies that only count. A one-line chat turn and a thousand-token reasoning trace are both ``one request'' to round-robin and to join-shortest-queue, yet they occupy a decode slot for durations that differ by two orders of magnitude. If a long request is sent to an instance that merely \emph{looks} idle, its SLO and the SLOs of everything queued behind it are missed.

The information needed to do better is available at admission and mostly free: the prompt length is exact, the output length can be estimated \cite{zheng2023response,fu2024ranking}, cache occupancy per instance is one metrics poll away, and the SLO class arrives with the request. Our question is whether a router that prices placements with these four signals beats counting-based placement on real hardware, by how much, where the advantage holds, and which signal contributes most.

We answer these questions with measurements on hardware. Our contributions are as follows:
\begin{enumerate}
\item \textbf{A marginal-cost learned router} that scores every candidate instance by the completion time this request would experience there, combining seconds-of-backlog, the request's own priced work, post-admission cache pressure, and SLO tightness (\S\ref{sec:method}).
\item \textbf{An end-to-end validation harness for routing policies on real disaggregated clusters}: a router proxy that drives unmodified vLLM engines over NIXL KV transfer, with per-request journaling, correctness gates (token-count contracts, cross-GPU greedy equivalence), per-run cache isolation, and per-workload saturation search (\S\ref{sec:setup}).
\item \textbf{A measured answer with boundaries} (\S\ref{sec:results}): the highest mean goodput and the lowest variance on mixed traffic; a $+4.5$-point contribution from hardware calibration alone; predicted output length identified as the load-bearing feature; robustness to 150\% predictor error; and two failure regions, narrow pools and extreme scarcity.
\end{enumerate}

\section{Background and Related Work}

\textbf{Disaggregation mechanisms.} DistServe \cite{zhong2024distserve} optimizes goodput per GPU under TTFT/TPOT targets by separating phases; Splitwise \cite{patel2024splitwise} splits across hardware pools with a cluster-level scheduler that joins shortest queues; Mooncake \cite{qin2024mooncake} organizes the cluster around a disaggregated KV cache and schedules with cache locality, load, and SLO state. The mechanisms are mature; the per-request placement policies remain simple, which is the gap this paper measures.

\textbf{Scheduling inside and across instances.} Continuous batching \cite{yu2022orca} and chunked prefill \cite{agrawal2024sarathi} decide execution order within an engine; Llumnix \cite{sun2024llumnix} migrates requests between instances after admission; AlpaServe \cite{li2023alpaserve} multiplexes models over shared GPUs. Our decision sits earlier, at the admission-time assignment across a prefill/decode fabric, which all of the above hold fixed.

\textbf{Output-length prediction.} Length estimators range from prompt-based regression \cite{zheng2023response} to learning-to-rank \cite{fu2024ranking}; TetriInfer \cite{hu2024tetriinfer} and LAPS \cite{cheng2026laps} schedule with predicted lengths. We treat the predictor as a swappable component and measure how much of its accuracy the router needs, which turns out to be very little.

\textbf{Robustness under bursts.} Steady-state goodput hides fragility; adaptive rescheduling exists because bursts cascade \cite{wang2025ares}. We evaluate at measured saturation points and report tails alongside means throughout. A recent survey covers the wider landscape \cite{li2024survey}.

\section{The Router}
\label{sec:method}

\begin{figure*}[t]
\centering
\includegraphics[width=0.96\textwidth]{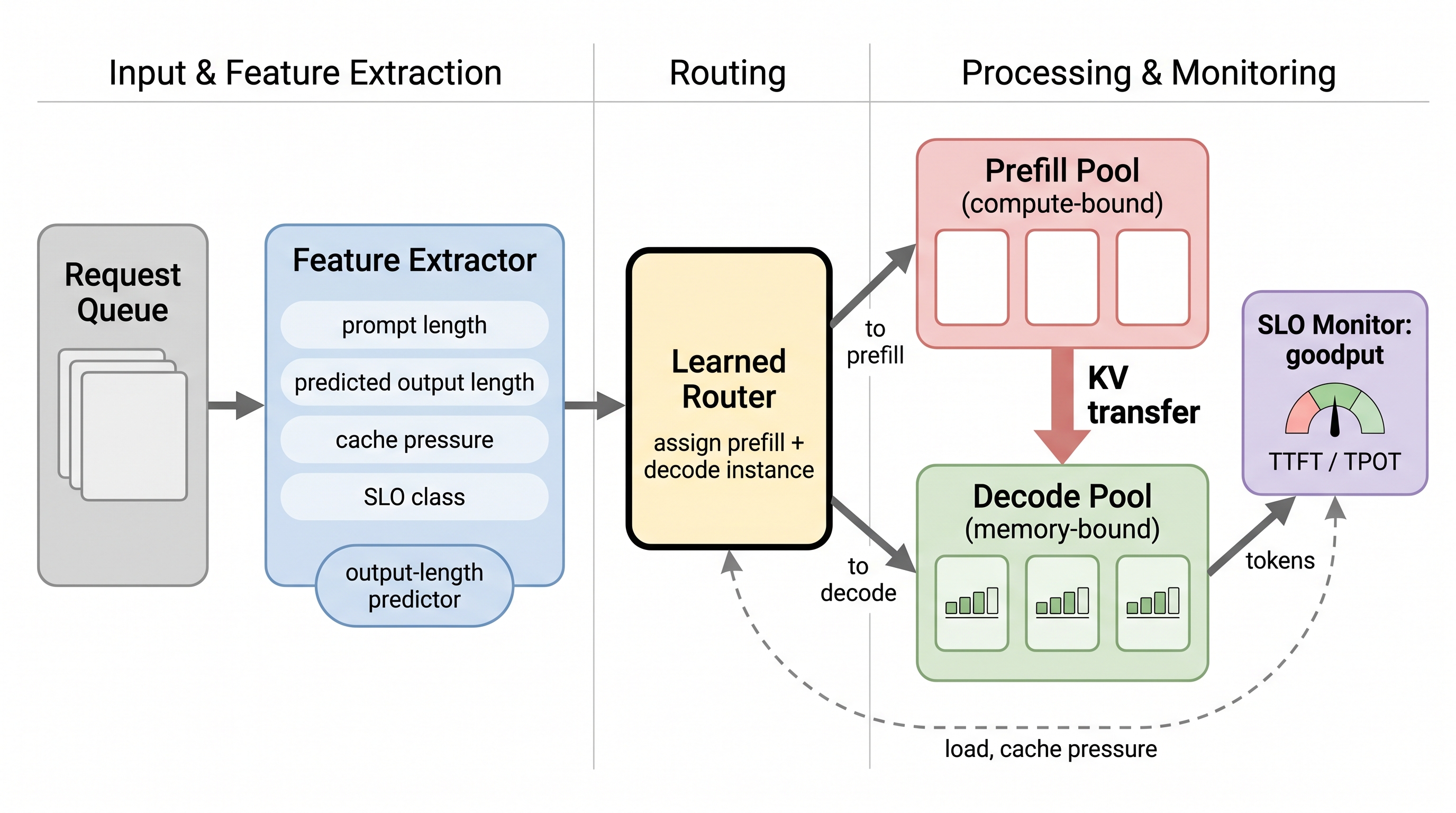}
\caption{Architecture. Requests are summarized at admission by prompt length, predicted output length, per-instance cache pressure, and SLO class; the router selects a prefill and a decode instance; the KV cache produced by prefill is transferred to the chosen decode instance; SLO attainment and live load signals feed back to the router. Baselines replace only the selection rule.}
\label{fig:method}
\end{figure*}

\subsection{Marginal-cost scoring}
For a request $r$ with prompt length $p_r$, predicted output length $\hat{o}_r$, and SLO class $s_r$, the router scores every decode candidate $d$ by an estimate of the completion time $r$ would experience there:
\begin{align*}
\mathrm{cost}(d) = {}& w_b \cdot \mathrm{backlog}(d) \cdot \sigma(s_r) \\
&+ w_s \cdot \hat{o}_r \, \tau_{\mathrm{dec}} \big(1 + \rho \cdot \mathrm{press}^{+}(d)\big) \cdot \sigma(s_r) \\
&+ w_q \cdot \mathrm{queue}(d),
\end{align*}
where $\mathrm{backlog}(d)$ is seconds of work already committed to $d$, $\tau_{\mathrm{dec}}$ is the hardware's per-token decode time, $\mathrm{press}^{+}(d)$ is cache pressure \emph{after} admitting $r$, and $\sigma(s_r)$ amplifies load terms for tight-SLO requests. Prefill candidates are scored analogously with $p_r \tau_{\mathrm{pre}}$. The request goes to the argmin of each pool. Algorithm~\ref{alg:route} summarizes.

\begin{algorithm}[t]
\caption{Learned routing (per request admission)}
\label{alg:route}
\begin{algorithmic}
\STATE \textbf{Input:} request $r$, prefill pool, decode pool
\STATE $\hat{o}_r \gets \text{predict\_output\_length}(p_r)$
\STATE $i_{\text{pre}} \gets \arg\min_{p} \mathrm{cost}_{\mathrm{pre}}(p; p_r, s_r)$
\STATE $i_{\text{dec}} \gets \arg\min_{d} \mathrm{cost}_{\mathrm{dec}}(d; \hat{o}_r, p_r, s_r)$
\STATE prefill on $i_{\text{pre}}$; transfer KV; decode on $i_{\text{dec}}$; record outcomes
\end{algorithmic}
\end{algorithm}

\subsection{Calibration is part of the method}
\label{sec:calibration}
The scorer mixes two kinds of quantities: $\mathrm{backlog}(d)$ is observed in real seconds, while the request's own work is \emph{priced} using $\tau_{\mathrm{dec}}$ and $\tau_{\mathrm{pre}}$. If those constants do not match the deployment hardware, the two terms are expressed in different units, the size-aware terms are effectively silenced, and the scorer degenerates into queue counting. We therefore define the router as calibrated by construction. A five-minute benchmark of the target cluster (prefill latency curve, decode rate under concurrency) sets $\tau_{\mathrm{pre}}$, $\tau_{\mathrm{dec}}$, and the SLO targets before any routing. Section~\ref{sec:res-calibration} measures what skipping this step costs; on our testbed the simulator-derived $\tau_{\mathrm{dec}}$ was $7.4\times$ smaller than the measured value.

\subsection{Output-length predictor}
The predictor supplies $\hat{o}_r$ and is deliberately simple. Because its quality bounds nothing else in the system, we evaluate the router under injected predictor noise up to 150\% relative error (\S\ref{sec:res-noise}).

\subsection{Development in simulation, evaluation on hardware}
The policy and features were developed in a deterministic discrete-event simulator with explicit pools, KV transfer, and SLO accounting. The simulator was used to generate hypotheses; none of the paper's numbers come from it. All results in \S\ref{sec:results} are measured on hardware, and the divergences we observed between simulated and measured behavior, in service model, in constants, and in which features matter, are themselves reported (\S\ref{sec:discussion}).

\section{Measurement Methodology}
\label{sec:setup}

\subsection{Testbed}
Eight NVIDIA A40 GPUs on one node, each running an unmodified vLLM 0.12 engine serving Qwen2.5-3B (BF16, TP=1), with KV caches moved between prefill and decode engines by NixlConnector. Each engine's KV budget is fixed at 60{,}000 tokens. The primary topology is 2 prefill + 4 decode instances; explicit pools of 3 to 8 GPUs serve the frontier and width studies. A FastAPI router proxy holds the policies (imported verbatim from the simulator code base), maintains live signals per instance (an in-flight ledger yielding seconds-of-backlog; cache occupancy polled every 50\,ms), executes the two-step disaggregated inference, and journals every request.

\subsection{Protocol}
Three properties separate this harness from a benchmark script. \emph{Correctness gates before any measurement:} an exact token-count contract per request; bitwise greedy-output equivalence across decode GPUs; and warmup of every prefill$\times$decode pair (the first transfer across a cold NIXL pair legitimately misbehaves). \emph{Isolation between runs:} every engine's prefix cache is reset before every run, after we observed identical configurations measuring 20+ points apart from session ordering. \emph{Saturation search per workload:} each workload family is driven at the arrival rate where round-robin lands mid-collapse, found by probing, because routing policies only separate under contention. SLO targets (TTFT 175\,ms / TPOT 39.7\,ms tight; $3\times$ loose) derive from the same calibration as the router constants. Every run is journaled per request and resumable. The headline experiments use three arrival traces.

\subsection{Workloads and baselines}
Four workload families stress different resources (Table~\ref{tab:data}). The baselines are round-robin, least-loaded (JSQ), and a predicted-length threshold heuristic.

\begin{table}[t]
\caption{Workloads and what each stresses.}
\label{tab:data}
\centering
\footnotesize
\setlength{\tabcolsep}{4pt}
\begin{tabularx}{\columnwidth}{@{}l>{\raggedright\arraybackslash}X>{\raggedright\arraybackslash}X@{}}
\toprule
Workload & Length profile & Stresses \\
\midrule
Chat & short in, short out & admission rate \\
Document & up to 4k in, short out & prefill compute, KV transfer \\
Reasoning & short in, long out & decode occupancy \\
Mixed bursty & heterogeneous, bursty & the routing decision itself \\
\bottomrule
\end{tabularx}
\end{table}

\section{Results}
\label{sec:results}

All figures and tables in this section are measured on the testbed of \S\ref{sec:setup}. Table~\ref{tab:main} rows average three arrival traces.

\begin{table}[t]
\caption{Mixed bursty traffic, three traces, 2 prefill + 4 decode. The learned router is best on the mean and has the smallest spread across traces. It beats round-robin and the heuristic on every trace; least-loaded edges it by $0.003$ on one of the three.}
\label{tab:main}
\centering
\footnotesize
\setlength{\tabcolsep}{4pt}
\begin{tabular}{@{}lccc@{}}
\toprule
Policy & Goodput & \begin{tabular}[b]{@{}c@{}}P99 TTFT\\(s)\end{tabular} & \begin{tabular}[b]{@{}c@{}}P99 TPOT\\(s)\end{tabular} \\
\midrule
Round-robin & 0.836 & 0.337 & 0.018 \\
Least-loaded (JSQ) & 0.835 & 0.346 & 0.018 \\
Heuristic length threshold & 0.847 & 0.318 & 0.018 \\
Learned router (ours) & \textbf{0.864} & 0.321 & 0.018 \\
\bottomrule
\end{tabular}
\end{table}

\subsection{Goodput under mixed bursty load}
Across the three traces the learned router posts 0.860/0.877/0.855, the highest mean and the smallest trace-to-trace spread, and on the single trace where a baseline edges it the margin is $0.003$. Least-loaded illustrates the cost of size-blindness, averaging respectably while dropping to 0.795 on its worst trace. Variance matters operationally, because SLAs are written against worst-case behavior, and the size-aware policy is the only one here with no weak trace.

\subsection{What calibration is worth}
\label{sec:res-calibration}
Running the identical learned policy with simulator-derived cost constants, on the same three traces, yields 0.819 mean goodput with 0.42 to 0.52\,s P99 TTFT, versus 0.864 and 0.25 to 0.30\,s calibrated. The calibration step alone is worth \textbf{4.5 goodput points and roughly 40\% of the tail advantage} (\S\ref{sec:calibration}). Per-request journals show the mechanism: with mispriced constants the router balances request \emph{counts} almost perfectly while letting token load skew worse than JSQ, i.e., it degenerates into the baseline it is supposed to beat. Any learned scheduler whose cost model is tuned in simulation inherits this failure mode silently.

\subsection{Effect of workload heterogeneity}
\begin{figure}[t]
\centering
\includegraphics[width=\columnwidth]{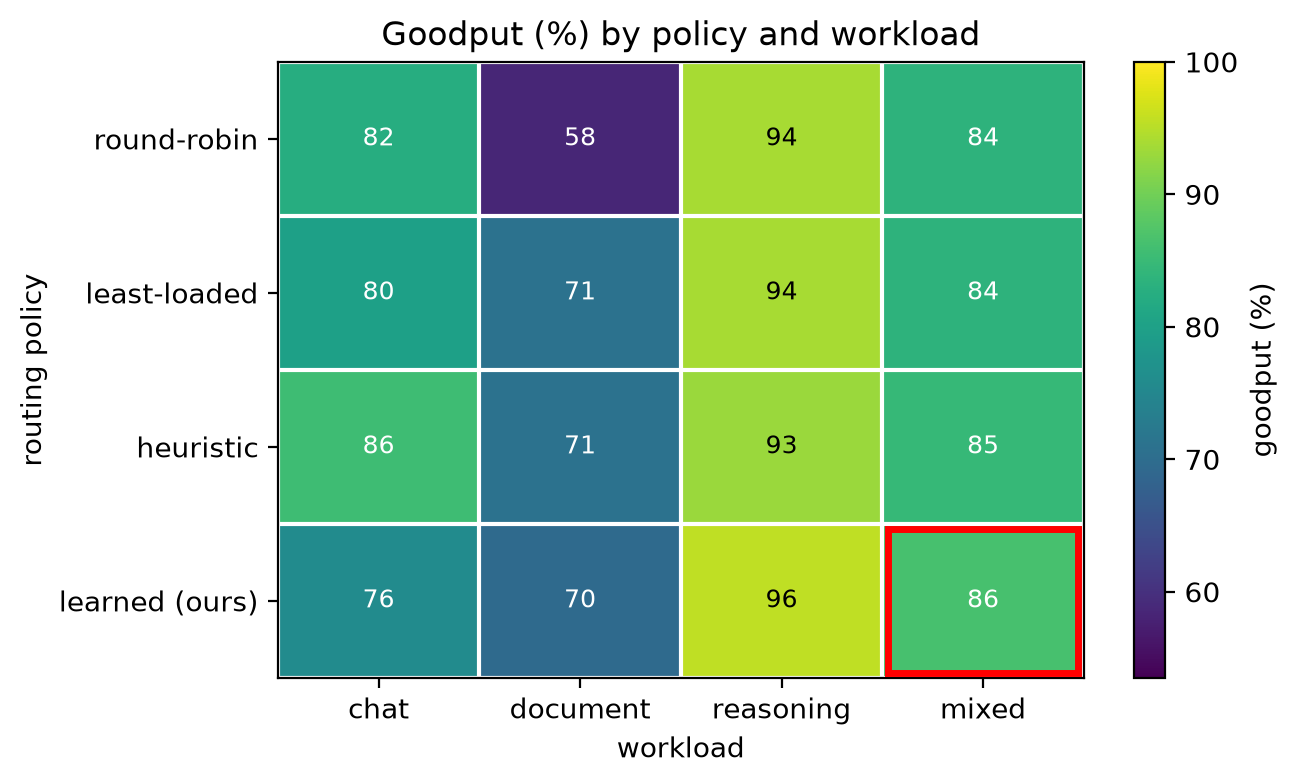}
\caption{Goodput by policy and workload, each workload at its own measured saturation point. Boxed cell: best per workload.}
\label{fig:policygrid}
\end{figure}

Figure~\ref{fig:policygrid} summarizes which policy suits which workload. The learned router wins where requests differ: mixed bursty traffic (0.864) and long-output reasoning (0.955). It is beaten on homogeneous chat (heuristic 0.855 vs.\ 0.760), where a queue count is close to a sufficient statistic and the additional signals add only noise. Round-robin's failure mode is concentrated and severe. On long-prompt documents it collapses to 0.585 with 1.1\,s tails while every load-aware policy holds ${\geq}0.695$.

\subsection{Effect of pool width}
We repeated the headline experiment at decode widths 3, 4, and 6, three traces each. At width 3 the learned router ties JSQ (0.816 vs.\ 0.822), at width 4 it leads outright (0.864 vs.\ 0.835), and at width 6, where this operating point leaves headroom, policies reconverge (0.858 vs.\ 0.860 for round-robin). The pattern matches the classical load-balancing result that the value of informed placement grows with the number of alternatives, and three instances offer too few for pricing to beat counting.

\subsection{Goodput versus provisioned GPUs}
\begin{figure}[t]
\centering
\includegraphics[width=\columnwidth]{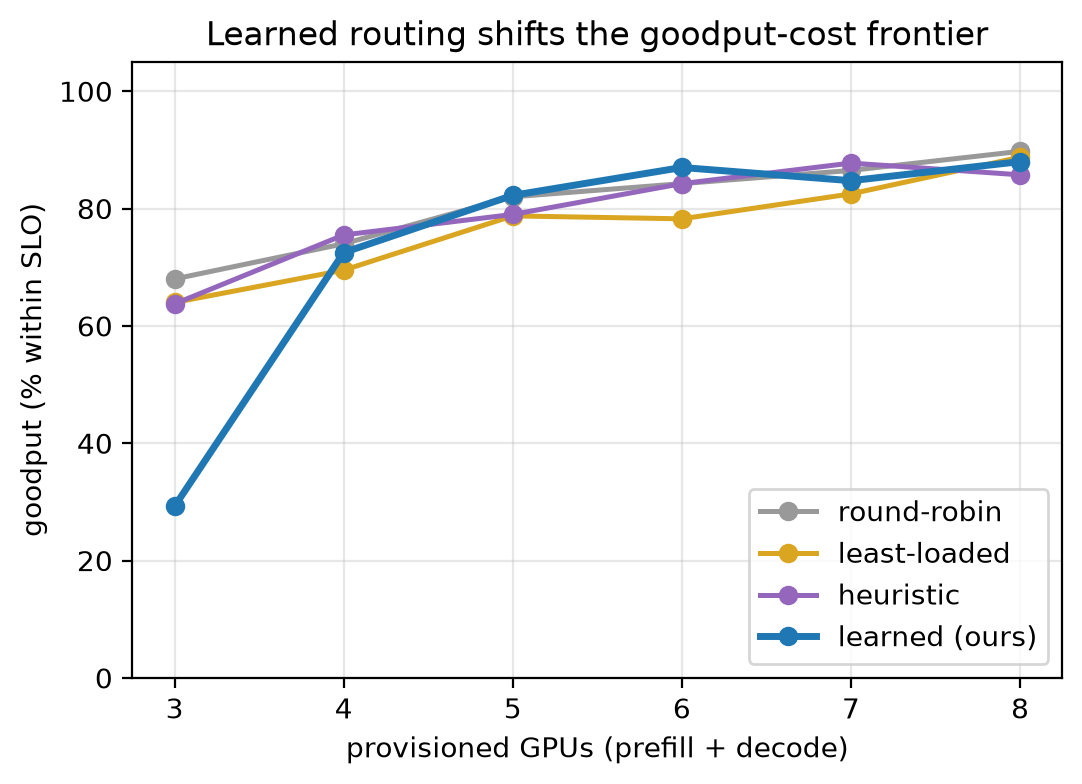}
\caption{Goodput versus provisioned GPUs across pool shapes. The learned router leads the deployable middle; the starved left end is its measured failure region.}
\label{fig:pareto}
\end{figure}

Figure~\ref{fig:pareto} sweeps pool shapes from 3 to 8 GPUs. The learned router leads the deployable middle of the frontier, best at 5 GPUs (0.823) and 6 GPUs (0.870), and reaches at 6 GPUs the goodput round-robin needs 7 for: \textbf{the same service with ${\sim}14\%$ fewer GPUs}. At the overprovisioned right end policies interleave within noise, as expected when every instance has headroom. At the starved left end (1 prefill + 2 decode) the learned router collapses to 0.292 while round-robin survives at 0.680; greedy cost minimization concentrates load on whichever instance momentarily prices cheapest, and when every instance is overloaded, spreading blindly is the better strategy. We reproduced this inversion four times, including with calibrated constants, so it is intrinsic. A sampled variant (score two random candidates) is the standard mitigation, which we leave to future work.

\subsection{Load and tails}
\begin{figure}[t]
\centering
\includegraphics[width=\columnwidth]{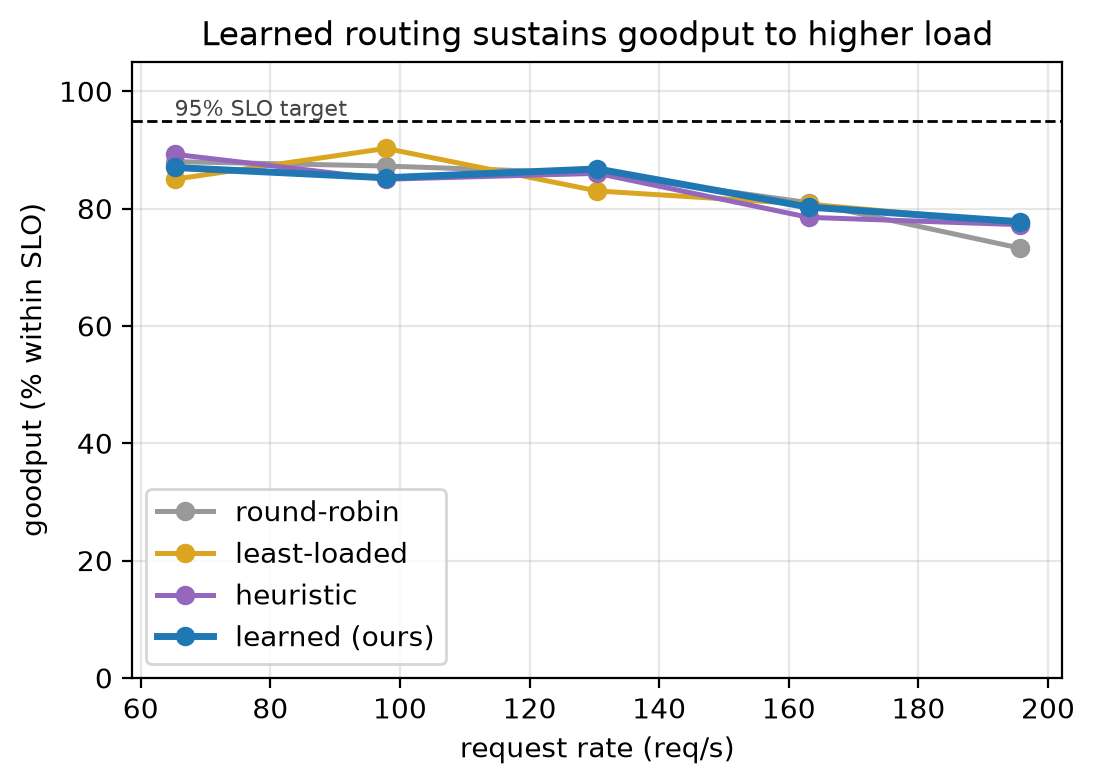}
\caption{Goodput versus request rate per policy.}
\label{fig:goodput}
\end{figure}

\begin{figure}[t]
\centering
\includegraphics[width=\columnwidth]{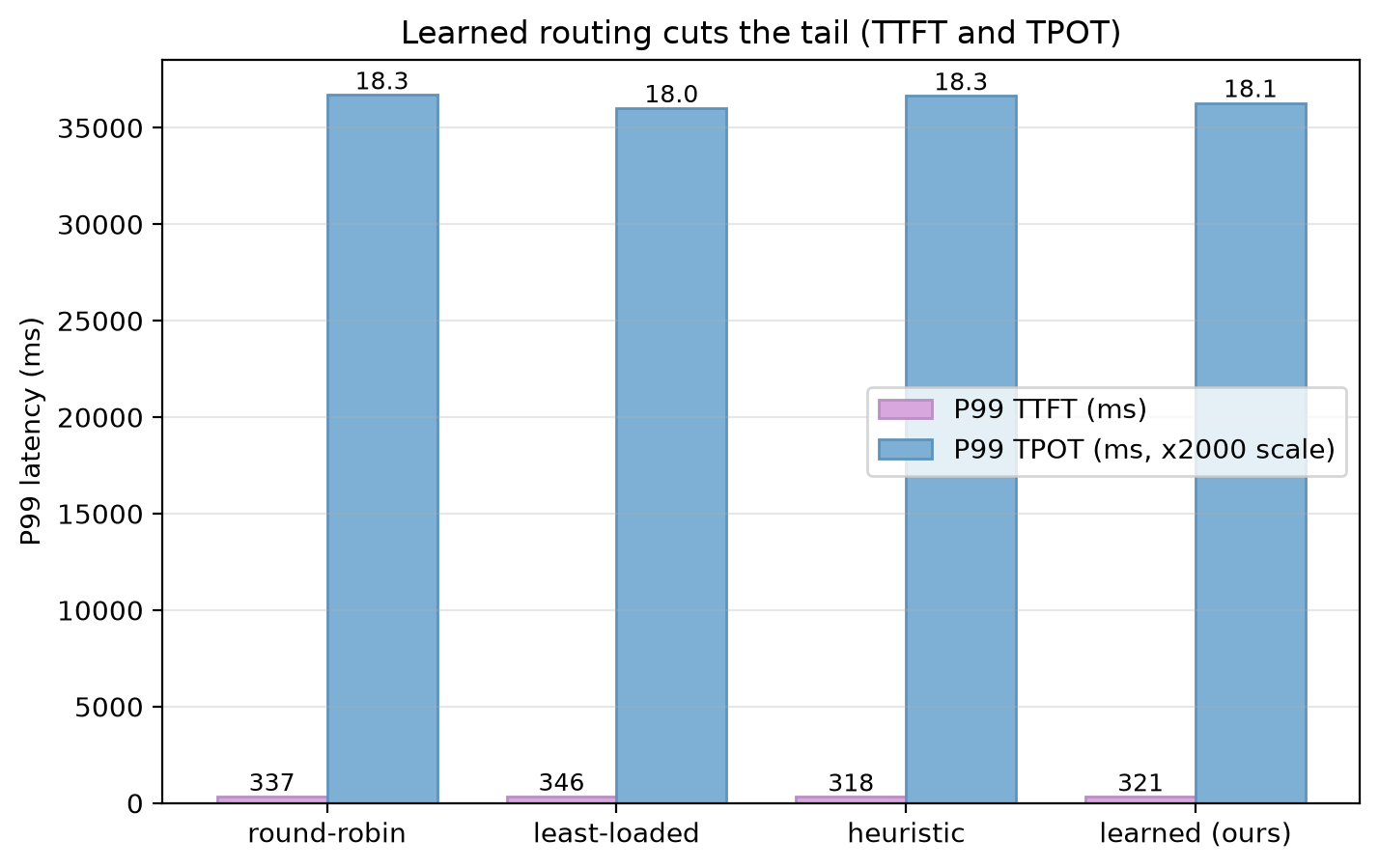}
\caption{P99 TTFT and TPOT per policy at the operating point.}
\label{fig:latency}
\end{figure}

\begin{figure}[t]
\centering
\includegraphics[width=\columnwidth]{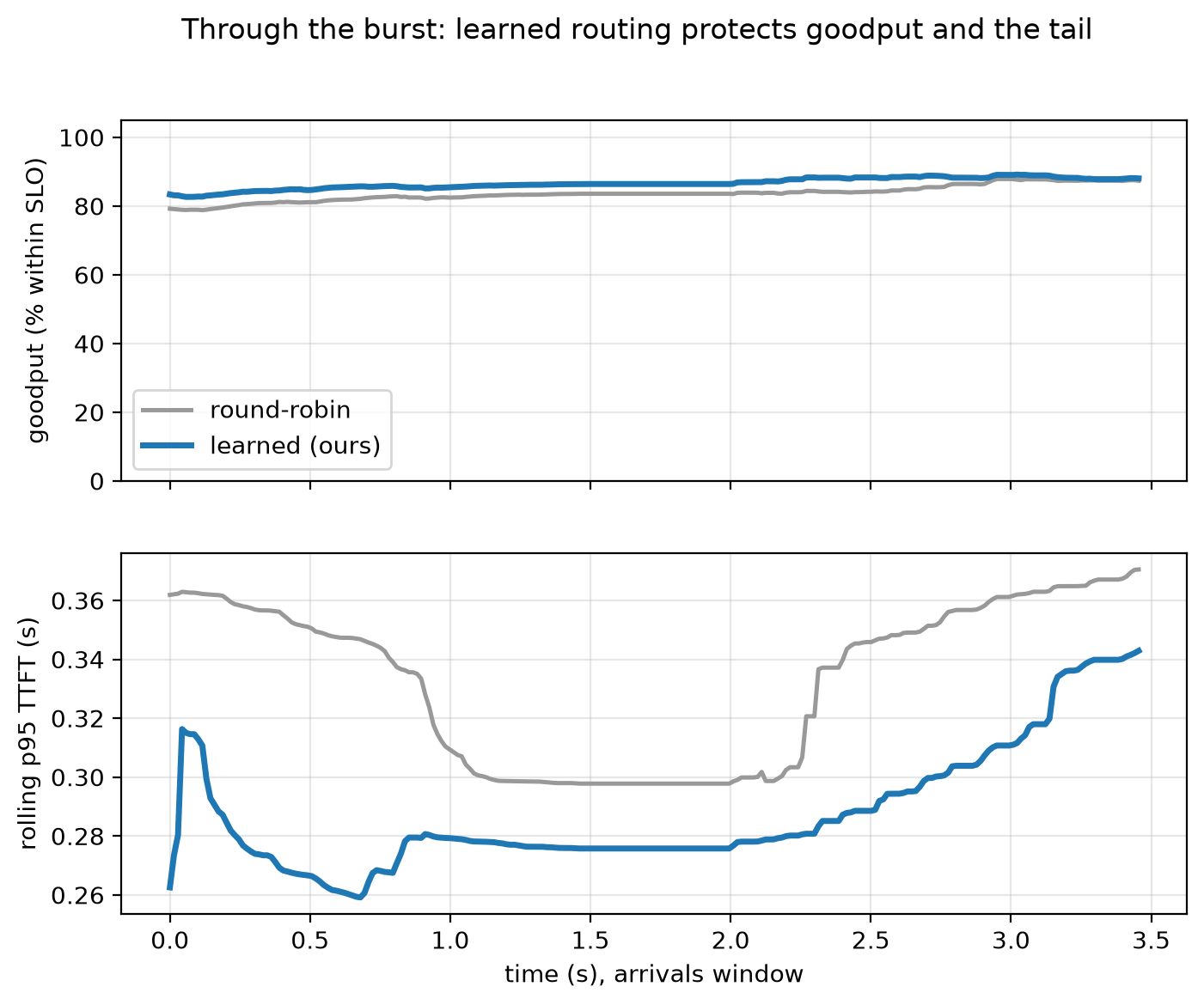}
\caption{Rolling goodput (top) and rolling P95 TTFT (bottom) through the burst train, learned versus round-robin.}
\label{fig:burst}
\end{figure}

Under a rising request rate (Fig.~\ref{fig:goodput}), round-robin degrades from 0.880 to 0.733 with P99 TTFT stretching from 0.22\,s to 0.56\,s; policies separate progressively with pressure. At the operating point (Fig.~\ref{fig:latency}) the learned router's P99 TTFT (0.321\,s) beats round-robin (0.337\,s) and JSQ (0.346\,s). Through a sustained burst train (Fig.~\ref{fig:burst}) it holds goodput above round-robin throughout and keeps rolling P95 TTFT roughly 20\% lower (0.28\,s vs.\ 0.30 to 0.37\,s). One measurement note: on a continuously batching cluster the aggregate in-flight count under overload is set by arrivals minus capacity regardless of policy, so queue depth carries no routing signal; the tail is where policies differ, and the burst figure reports it accordingly.

\subsection{Which feature does the work}
\label{sec:res-ablation}
\begin{figure}[t]
\centering
\includegraphics[width=\columnwidth]{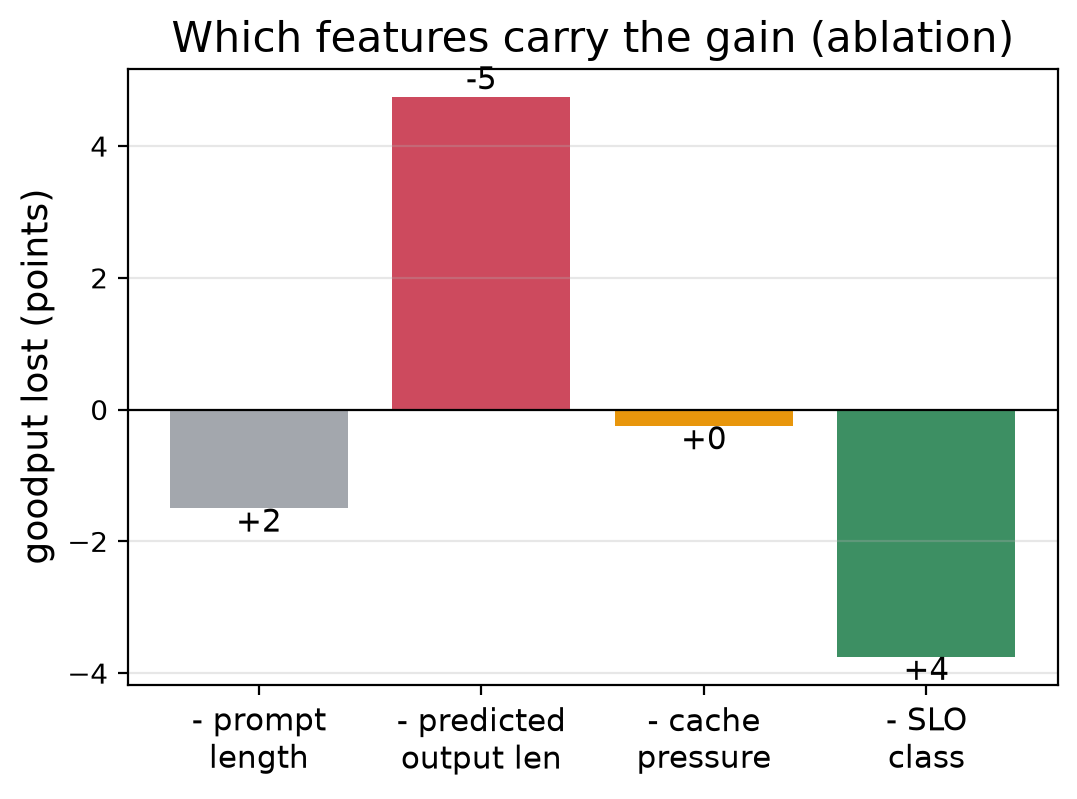}
\caption{Goodput lost when each feature is removed (signed; a below-zero bar means removal helped at that operating point).}
\label{fig:ablation}
\end{figure}

We removed features one at a time at two operating points (Fig.~\ref{fig:ablation}). \textbf{Predicted output length is the load-bearing feature}, costing 3.2 points when removed under steady load and 4.7 under saturation, the largest loss in both regimes. Cache pressure contributes nothing measurable here because the 60k-token KV budget never saturates; this does not show that the feature is unhelpful, only that its regime was not reached here. The SLO term helps at moderate load ($-1.7$) and mildly backfires under deep saturation ($+3.7$), a regime dependence that reached $-12$ points with uncalibrated constants and stays within a few points with calibrated ones.

\subsection{How good the predictor needs to be}
\label{sec:res-noise}
\begin{figure}[t]
\centering
\includegraphics[width=\columnwidth]{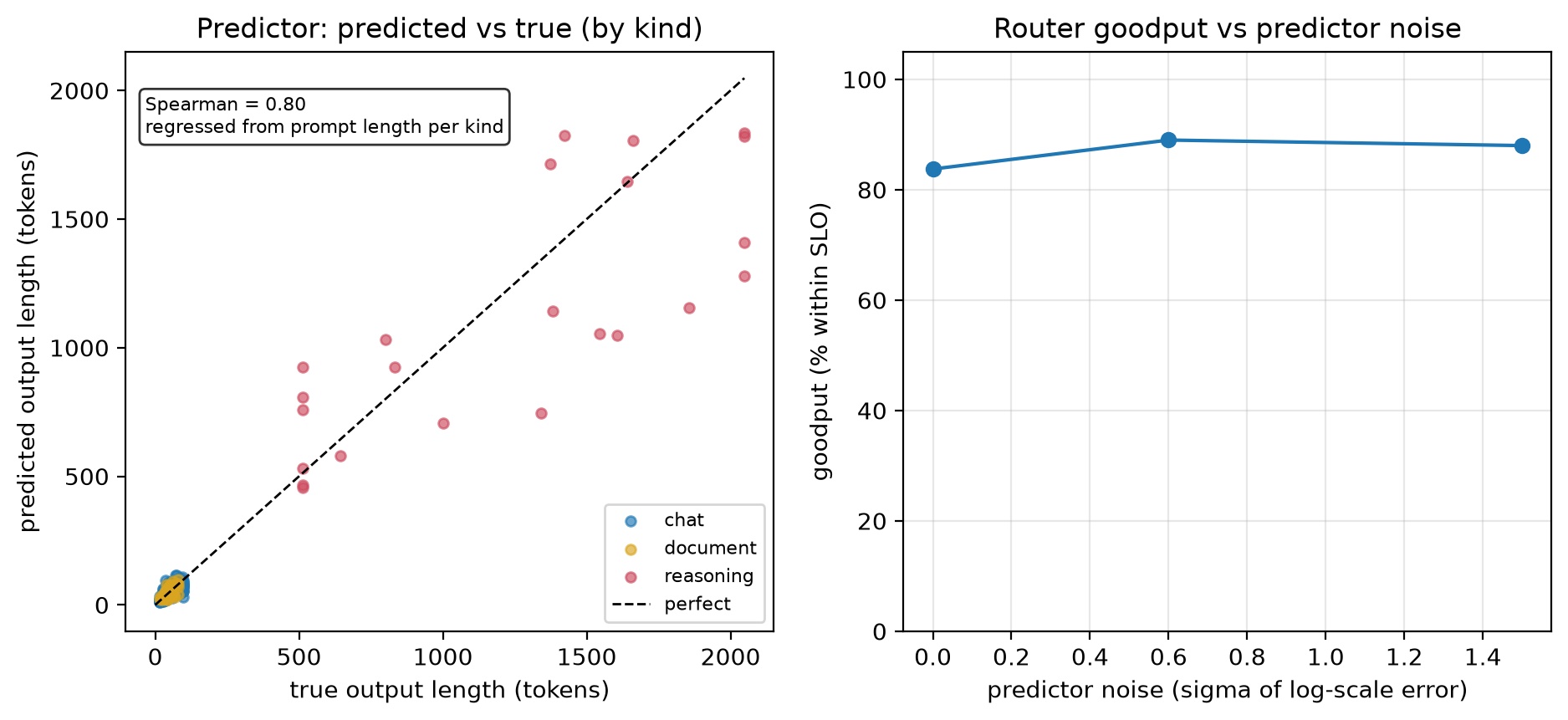}
\caption{Router goodput as predictor error is injected; degradation is flat to 150\% error.}
\label{fig:lengthpred}
\end{figure}

Injecting relative error into the output-length predictor from 0 to 150\% moves goodput from 0.838 to 0.880, i.e., within trace noise, and the router still beats the heuristic at the worst injected error (Fig.~\ref{fig:lengthpred}). This is convenient for deployment, since a magnitude estimate suffices in place of an accurate model, consistent with the ablation's finding that \emph{having} a length signal matters far more than its precision.

\section{Discussion and Limitations}
\label{sec:discussion}

\textbf{Simulation predicts structure, not magnitudes.} Our simulator's cost constants were $7$ to $13\times$ off the measured hardware, and its serial-service model overstates head-of-line blocking that continuous batching softens. To quantify the gap, we re-fitted the simulator to the measured A40 physics (decode $0.0018 \rightarrow 0.013237$\,s/token, prefill base $0.005 \rightarrow 0.0667$\,s, KV bytes/token $2048 \rightarrow 36864$) and re-anchored each workload's arrival rate on the \emph{measured round-robin} goodput, so that its verdict on the other three policies is a prediction rather than a fit. Across the resulting $4 \times 4$ grid the simulator reproduces goodput to a mean absolute residual of $0.068$, but it does \emph{not} reproduce the measured winner on any workload. We also implemented continuous batching in both stages, since its absence was the obvious suspect. The measured decode curve is flat (a batch of eight costs each request $2.9\%$ more per token), whereas a serial model makes the eighth request wait $8\times$. Batching corrected the physics decisively: p99 TTFT went from $1$ to $63\times$ overstated to $0.5$ to $1.4\times$ on three of four workloads, and matching the measured round-robin goodput went from requiring an arrival rate $1700\times$ below the hardware's to requiring nearly the hardware's own rate ($24.6$ against $24.0$). It did not improve the ranking at all. Mean residual moved $0.067 \rightarrow 0.068$, and with batching all four policies fall within $0.01$ of each other on bursty where the hardware separated them by $0.03$. Our simulator's earlier apparent skill was therefore an artefact of serialisation, which inflated the value of knowing output length and made the learned router appear correct for the wrong reason. What a serial-plus-batching queueing model still lacks is whatever makes placement \emph{consequential} at a fixed goodput level on real hardware: intra-batch interference between long and short sequences, chunked-prefill contention, and KV fragmentation. The routing guide of \S\ref{sec:results} is therefore a hardware result, and we do not claim it is recoverable from simulation.

\textbf{Why calibration is invisible in simulation.} The $+4.5$-point calibration effect of \S\ref{sec:res-calibration} does not reproduce in our simulator at all, and the reason is structural rather than numerical. The simulator sets each instance's backlog signal from the cluster's own clock, so the router receives its \emph{true} remaining work; the deployed proxy must estimate that backlog from the router's own assumed constants, which is where a $7.4\times$ error does its damage. Restoring the proxy's estimator inside the simulator brings the mechanism back qualitatively (a miscalibrated router's p99 TTFT degrades $0.899 \rightarrow 3.461$\,s), but it recovers almost none of the goodput effect ($+0.1$ versus $+4.5$ points), which we attribute to the batch-level interference a serial-service model cannot express. The broader caution is that a simulator can hand a policy information no deployment will have, and that this advantage stays invisible until the policy meets hardware.

\textbf{Boundaries observed.} The router's advantage requires alternatives to price (width ${\geq}4$ here), heterogeneous traffic, and a cluster that is contended but not saturated beyond capacity. Under extreme scarcity, greedy argmin placement concentrates load on the momentarily cheapest instance, and production deployments should compose the scorer with power-of-two sampling or an admission guard. Cache pressure was inert at our KV budget; larger contexts or smaller budgets would exercise it.

\textbf{Scope.} One silicon per experiment set, single-node NIXL transfer (cross-node RDMA costs would add a transfer term the scorer already accommodates), a 3B model, and a scoring function with hand-set weights whose constants, not weights, were fitted; the production systems converging on this design (KV-aware cost routers in NVIDIA Dynamo, Mooncake's Conductor, llm-d's scorer plugins) suggest the pattern generalizes, with cache-locality terms as the natural next feature.

\section{Conclusion}
On a real disaggregated vLLM/NIXL cluster, a marginal-cost router over four admission-time features delivers the highest mean goodput on mixed traffic with the lowest variance across traces, matches the service of blind routing with about 14\% fewer GPUs, and needs only a crude output-length estimate to do it. Knowing the hardware is an essential part of the method. Calibrating the cost model to measured per-token times is worth 4.5 goodput points by itself, and its absence silently reduces a learned router to the queue counting it was built to replace. The advantage is bounded by pool width, by traffic heterogeneity, and by a load-concentration failure under scarcity; we report those boundaries as measured results, since they and the gains together are what a deployer needs to know.

\section*{Acknowledgments}
The authors thank Vizuara AI Labs for mentorship and computational resources.


\begin{thebibliography}{99}
\bibitem{zhong2024distserve} Y. Zhong et al., ``DistServe: disaggregating prefill and decoding for goodput-optimized large language model serving,'' \textit{Proc. OSDI}, 2024. arXiv:2401.09670.
\bibitem{patel2024splitwise} P. Patel et al., ``Splitwise: efficient generative LLM inference using phase splitting,'' \textit{Proc. ISCA}, 2024. arXiv:2311.18677.
\bibitem{qin2024mooncake} R. Qin et al., ``Mooncake: a KVCache-centric disaggregated architecture for LLM serving,'' arXiv:2407.00079, 2024.
\bibitem{kwon2023vllm} W. Kwon et al., ``Efficient memory management for large language model serving with PagedAttention,'' \textit{Proc. SOSP}, 2023. arXiv:2309.06180.
\bibitem{agrawal2024sarathi} A. Agrawal et al., ``Taming throughput-latency tradeoff in LLM inference with Sarathi-Serve,'' \textit{Proc. OSDI}, 2024. arXiv:2403.02310.
\bibitem{sun2024llumnix} B. Sun et al., ``Llumnix: dynamic scheduling for large language model serving,'' \textit{Proc. OSDI}, 2024. arXiv:2406.03243.
\bibitem{yu2022orca} G. Yu et al., ``Orca: a distributed serving system for transformer-based generative models,'' \textit{Proc. OSDI}, 2022.
\bibitem{li2023alpaserve} Z. Li et al., ``AlpaServe: statistical multiplexing with model parallelism for deep learning serving,'' \textit{Proc. OSDI}, 2023. arXiv:2302.11665.
\bibitem{zheng2023response} Z. Zheng et al., ``Response length perception and sequence scheduling: an LLM-empowered LLM inference pipeline,'' \textit{Proc. NeurIPS}, 2023. arXiv:2305.13144.
\bibitem{fu2024ranking} Y. Fu et al., ``Efficient LLM scheduling by learning to rank,'' \textit{Proc. NeurIPS}, 2024. arXiv:2408.15792.
\bibitem{hu2024tetriinfer} C. Hu et al., ``Inference without interference: disaggregate LLM inference for mixed downstream workloads,'' arXiv:2401.11181, 2024.
\bibitem{cheng2026laps} K. Cheng et al., ``LAPS: a length-aware-prefill LLM serving system,'' arXiv:2601.11589, 2026.
\bibitem{wang2025ares} Y. Wang et al., ``ARES: adaptive rescheduling in prefill-decode disaggregated LLM inference,'' arXiv:2510.13668, 2025.
\bibitem{li2024survey} B. Li et al., ``LLM inference serving: survey of recent advances and opportunities,'' arXiv:2407.12391, 2024.
\bibitem{dao2022flashattention} T. Dao et al., ``FlashAttention: fast and memory-efficient exact attention with IO-awareness,'' \textit{Proc. NeurIPS}, 2022. arXiv:2205.14135.
\bibitem{zhang2023h2o} Z. Zhang et al., ``H2O: heavy-hitter oracle for efficient generative inference of large language models,'' \textit{Proc. NeurIPS}, 2023. arXiv:2306.14048.
\bibitem{agrawal2024vidur} A. Agrawal et al., ``Vidur: a large-scale simulation framework for LLM inference,'' \textit{Proc. MLSys}, 2024. arXiv:2405.05465.
\bibitem{dubey2024llama3} A. Dubey et al., ``The Llama 3 herd of models,'' arXiv:2407.21783, 2024.
\bibitem{yang2024qwen25} A. Yang et al., ``Qwen2.5 technical report,'' arXiv:2412.15115, 2024.
\bibitem{stojkovic2024dynamollm} J. Stojkovic et al., ``DynamoLLM: designing LLM inference clusters for performance and energy efficiency,'' arXiv:2408.00741, 2024.
\bibitem{lin2024infinite} C. Lin et al., ``Infinite-LLM: efficient LLM service for long context with distributed attention,'' arXiv:2401.02669, 2024.
\bibitem{reddi2020mlperf} V. J. Reddi et al., ``MLPerf inference benchmark,'' \textit{Proc. ISCA}, 2020. arXiv:1911.02549.
\end{thebibliography}
\end{document}